\documentclass[journal]{IEEEtran}
\usepackage{amsmath,amsfonts}
\usepackage{algorithmic}
\usepackage{algorithm}
\usepackage{array}
\usepackage[caption=false,font=normalsize,labelfont=sf,textfont=sf]{subfig}
\usepackage{textcomp}
\usepackage{stfloats}
\usepackage{url}
\usepackage{verbatim}
\usepackage{graphicx}
\usepackage{cite}
\usepackage{booktabs}
\usepackage{ulem}
\usepackage{tabularx}
\usepackage{multirow}

\begin{document}

\title{MFSR: Multi-fractal Feature for Super-resolution Reconstruction with Fine Details Recovery}

\author{Lianping Yang, Peng Jiao, Jinshan Pan, Hegui Zhu, Su Guo~\IEEEmembership{}
\thanks{Lianping Yang and Peng Jiao contributed equally to this paper. This work is supported by the National Natural Science Foundation of China (No. 62472080) Corresponding author: Lianping Yang (Email: yanglp@mail.neu.edu.cn).

Lianping Yang, Peng Jiao and Hegui Zhu are with the College of Sciences, Northeastern University, Shengyang 110819, China

Lianping Yang, Hegui Zhu are also with Key Laboratory of Differential Equations and Their Applications, Northeastern University, Liaoning Provincial Department of Education

Jinshan Pan is with School of Computer Science and Engineering, Nanjing University of Science, Nanjing, China

Su Guo is with College of Renewable Energy, Hohai University, Nanjing 211000, China
}

}


\maketitle

\begin{abstract}
In the process of performing image super-resolution processing, the processing of complex localized information can have a significant impact on the quality of the image generated. Fractal features can capture the rich details of both micro and macro texture structures in an image. Therefore, we propose a diffusion model-based super-resolution method incorporating fractal features of low-resolution images, named MFSR. MFSR leverages these fractal features as reinforcement conditions in the denoising process of the diffusion model to ensure accurate recovery of texture information. MFSR employs convolution as a soft assignment to approximate the fractal features of low-resolution images. This approach is also used to approximate the density feature maps of these images. By using soft assignment, the spatial layout of the image is described hierarchically, encoding the self-similarity properties of the image at different scales. Different processing methods are applied to various types of features to enrich the information acquired by the model. In addition, a sub-denoiser is integrated in the denoising U-Net to reduce the noise in the feature maps during the up-sampling process in order to improve the quality of the generated images. Experiments conducted on various face and natural image datasets demonstrate that MFSR can generate higher quality images.
\end{abstract}

\begin{IEEEkeywords}
fractal, multi-fractal, super-resolution, diffusion model
\end{IEEEkeywords}

\section{Introduction}
\IEEEPARstart{S}{ingle} Image Super-Resolution (SISR) is a crucial task in the field of computer vision. This process involves inputting a low-resolution (LR) image and generating a corresponding high-resolution (HR) image, with the aim of restoring and enhancing the image's quality. Low-resolution images often lack the detailed information present in high-resolution images. Therefore, an essential aspect of super-resolution is the recovery and reconstruction of these details to achieve a higher quality output.

\begin{figure}[h]
    \centering
    \includegraphics[width=0.5\textwidth]{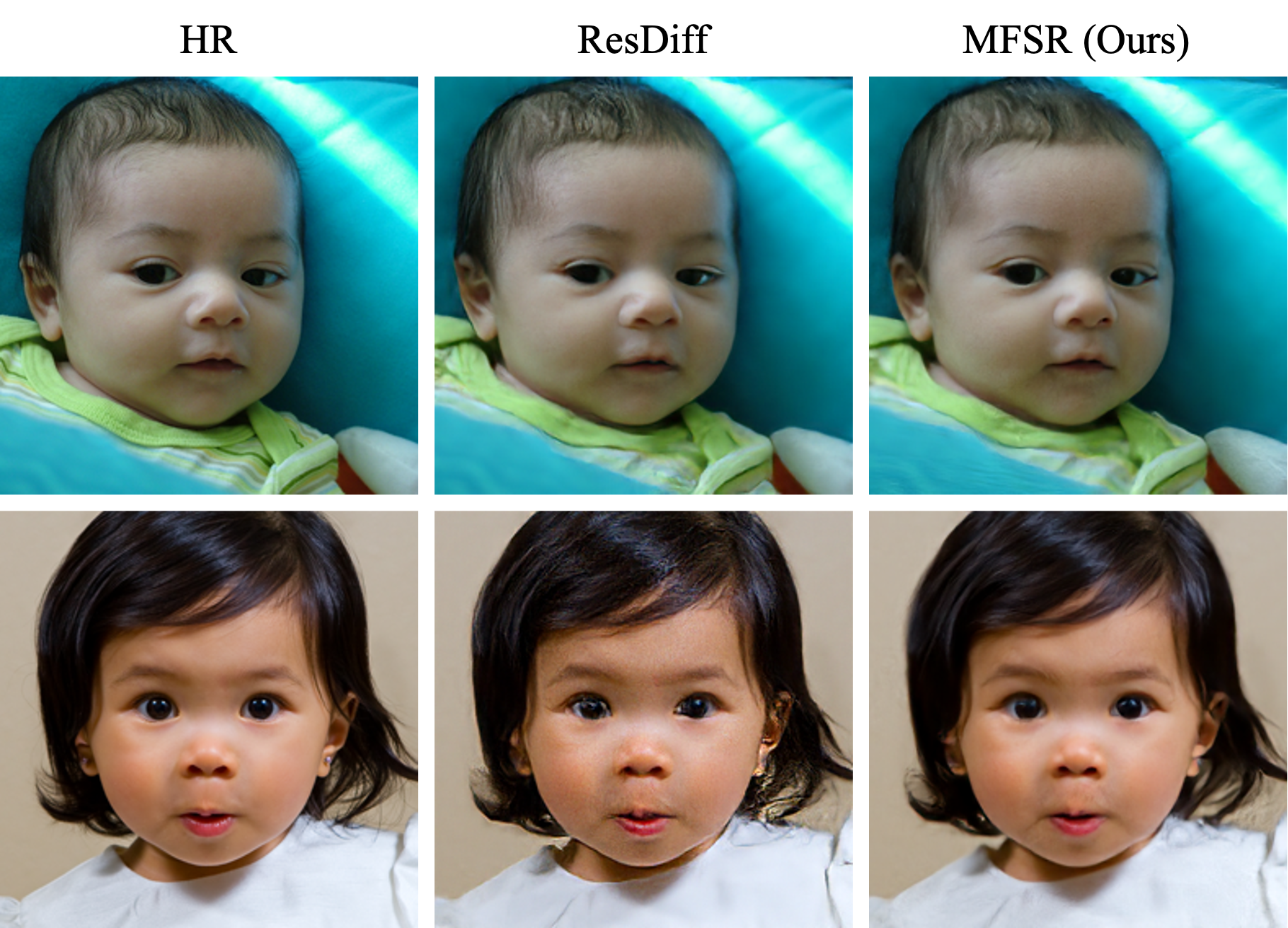}
    \caption{This is a demonstration of the results of MFSR's experiments on the FFHQ dataset. Each row represents a sample. The first column represents the input HR image, the second column represents the result of ResDiff, and the third column represents the result of MFSR (Ours). MFSR has better super-resolution effect on details than ResDiff.}
    \label{fig:1a}
\end{figure}

At this stage, many methods have been developed to address the problem of image super-resolution. Since the advent of deep neural networks, numerous CNN-based approaches have been utilized, employing a variety of neural network structures to process low-resolution (LR) images and generate high-resolution (HR) images\cite{super_resolution1}. Image super-resolution can also be viewed as a process of new image generation. Consequently, since the introduction of Generative Adversarial Networks (GANs)\cite{GAN1}, many generation-based super-resolution methods have been proposed. Techniques such as SRGAN\cite{srgan1} and ESGAN\cite{esrgan1} use GANs for super-resolution processing of images. However, due to inherent limitations of GANs, these methods can be prone to instability and may crash during training\cite{crash1}. Therefore, a more robust generative model is needed for the task of image super-resolution. Denoising Diffusion Probabilistic Model (DDPM) also has numerous applications in the field of image super-resolution, such as SR3\cite{sr3}, SRDiff\cite{srdiff}, ResDiff\cite{resdiff}, ResShift\cite{resshift} and so on. Stable Diffusion\cite{stablediffusion} demonstrates powerful generative capabilities in image generation. Further, many super-resolution methods based on this have been proposed, such as DiffBIR\cite{diffbir}, PASD\cite{pasd}. Although there are many models based on spreading models for super-resolution, most of them do not focus on texture information as well as self-similar features.

Image texture is extremely critical to visual perception and detail expression. They are indispensable in portraying details of object surfaces, shapes, and material differences. Multi-fractal feature has the unique advantage of accurately capturing multi-scale texture variations based on the self-similarity principle in texture, such as fabric images, which can present micro-fibers and macro-patterned textures. In terms of details, multi-scale analysis can focus on tiny details and integrate and reconstruct them after decomposition processing\cite{chongfenxing1}. In terms of self-similarity, it can utilize the self-similar structure of images at different scales to infer high-resolution information based on the relationship between the local area and the whole, to improve the resolution and maintain natural coherence. Therefore, multi-fractal feature is suitable for image super-resolution tasks.

This paper aims to propose a new method for super-resolution reconstruction to address the inadequacies of current diffusion model-based methods in recovering image texture information by multi-fractal.

\begin{figure}[h]
    \centering
    \includegraphics[width=0.3\textwidth]{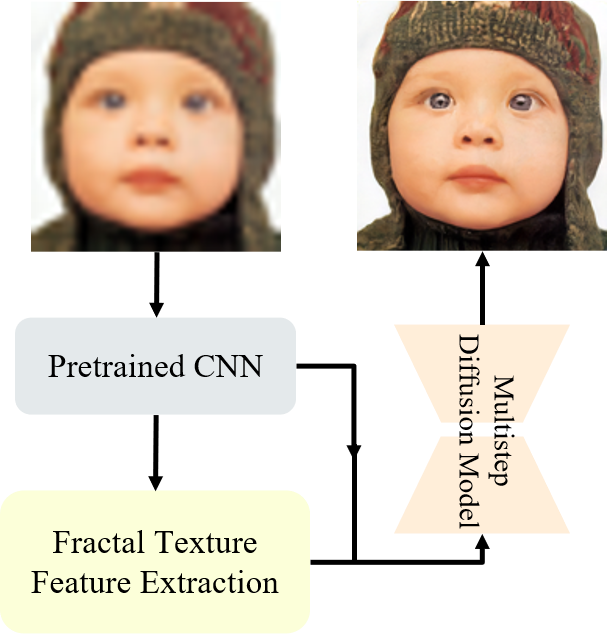}
    \caption{A brief flowchart of the MFSR. Texture features are extracted for LR images processed by pretrained CNNs. Splicing texture features with other features as model input.}
    \label{fig:15}
\end{figure}

Fractal features are capable of capturing rich information about both micro-texture and macro-texture structures in images, making them highly promising for image analysis. However, current super-resolution reconstruction methods often fail to fully utilize the advantages of fractal features in texture information description. Inspired by this, we propose MFSR, which use multi-fractal feature for super-resolution based on DDPM. The rough model diagram is shown in Fig\ref{fig:15}. MFSR uses a Multi-Fractal Feature Extraction Block (MFB), which utilizes convolution to approximate the multi-fractal features of low-resolution images and encodes these features at different scales. This approach encodes the self-similar properties of the image across various scales, enriching the information acquired by the model. Additionally, DDPM typically employs a U-Net\cite{unet} architecture as the denoising network. We use an attention-based denoiser. The feature maps converted to the frequency domain space are denoised so as to reduce the noise perturbation in the denoising process.


Experiments conducted on various datasets demonstrate that MFSR can generate fine-grained images with superior results, as illustrated in Figure \ref{fig:1a}.

Our contributions are as follows:

\begin{itemize}

\item The multi-fractal features of LR image are incorporated into the SR task in the form of texture prior to improve image quality. Experiments show that it can obtain better SR results.

\item By incorporating a sub-denoiser into the U-Net cascade, we reduce the impact of noise in the denoising process. Experiments show that it can obtain better SR results.

\item Experiment results show that our method can generate more refined SR images.

\end{itemize}

\section{Related Works}
Generative model-based image super-resolution methods can be primarily categorized into three types: Flow-based methods, GAN-based methods, and Diffusion-based methods. The following sections introduce each of these approaches. We will also present work on super-resolution tasks combined with fractal.
\subsection{Flow-based Methods}
A flow-based approach encodes the original image into latent space using a function $f$, and then samples from the latent space to recover the image via the inverse function $f^{-1}$. Lugmayr et al. proposed a flow-based super-resolution model capable of learning the conditional probability distribution of a given low-resolution image, utilizing only one loss function: negative log-likelihood. Notably, SRFlow\cite{srflow} outperforms many current GAN-based methods in face super-resolution scenarios. While flow-based methods align well with mathematical intuitions, performing reversible processing is more challenging for neural networks.
\subsection{GAN-based Methods}
After the introduction of GANs\cite{GAN1}, Ledig et al.\cite{srgan1} designed SRGAN, the first application of GANs for image super-resolution tasks, introducing a perceptual loss function to enhance image generation quality. Building on SRGAN, Kim et al.\cite{esrgan1} introduced RRDB and utilized features before activation for perceptual loss calculation, resulting in ESRGAN\cite{esrgan1}, which aimed for improved super-resolution outcomes. GAN-based super-resolution methods combine various loss functions, enabling the model to generate higher quality super-resolution images. However, the adversarial training approach of GANs often leads to pattern collapse during training\cite{crash1}.
\subsection{Diffusion-based Methods}
DDPM\cite{DDPM1} employs a stepwise denoising process capable of producing clear images. Saharia et al. proposed SR3,\cite{sr3} which combines low-resolution images with noise-laden feature maps in a denoiser, achieving strong performance across various super-resolution tasks. Li et al.\cite{srdiff} introduced SRDiff, which incorporates residual prediction throughout the framework. In this approach, the original image is encoded through an encoder for processing conditional images, leading to improved super-resolution results. Shang et al.\cite{resdiff} proposed ResDiff, which utilizes a simple CNN to recover the low-frequency components of the image, while DDPM predicts the residuals between the real image and the CNN-predicted image. Additionally, high-frequency information is introduced into the denoising network using Discrete Wavelet Transform (DWT) information. The ResShift proposed by Yue et al.\cite{resshift} realizes the conversion between high-resolution images and low-resolution images by shifting the residuals between them, thus greatly improving the conversion efficiency. The PASD network proposed by Yang et al.\cite{pasd} utilizes a high-level information extraction module to provide semantic signals for realistic image super-resolution and personalized stylization. The DiffBIR proposed by Lin et al.\cite{diffbir} balances the realism a priori inherent in the diffusion model and the fidelity requirements needed for the image restoration task. The strong generative power of the diffusion model makes the image super-resolution effect better.


\section{Preliminary}

\subsection{Multi-Fractal Analysis Methods}

Fractal describes a shape, pattern, or process that is self-similar over a range of spatial scales\cite{process1}. Many natural objects and phenomena exhibit self-similar or fractal properties: a structure is assumed to consist of parts that are similar to the whole. Natural textures carry much information about the fractal dimension.

To compute the fractal dimension, the number of non-empty boxes $N(\epsilon)$ is computed by covering the structure $S$ with boxes of size $\epsilon$ . When $\epsilon$ tends to zero, the limit value of $N(\epsilon)$ obeys a power law\cite{milv}, i.e., the fractal dimension is estimated as:
\begin{equation}\label{equ5}
D_f=-\lim _{\varepsilon \rightarrow 0} \frac{\ln (N(\varepsilon))}{\ln (\varepsilon)}.
\end{equation}

A wide variety of natural objects exhibit self-similarity. However, beyond a certain scale structures may no longer be fractal, and structures characterized by fractal dimensions that vary with the scale of the observation are known as multi-fractal\cite{chongfenxing1}.

The structure $S$ is partitioned into non-overlapping boxes $S_i$ of size $\epsilon$, each characterized by the measure $\mu(S_i)$. Thus, an equivalent parameter for the multi-fractal spectral analysis is defined as $\alpha_i=\frac{\ln \left(\mu\left(S_i\right)\right)}{\ln (\varepsilon)}.$ $\alpha_i$ is denoted as the coarse Holder index\cite{fenxing3} of the subset $S_i$. As $\epsilon$ tends to 0, the coarse Holder index tends to the limiting value $\alpha=\lim _{\varepsilon \rightarrow 0}\left(\alpha_i\right)$ at the observation point.


The parameter $\alpha$ describes the local regularity of the structure. In the whole structure $S$, there are usually many boxes or points with the same parameter $\alpha$. Therefore the process of finding a function $f(\alpha)$ on a subset characterized by $\alpha$. This function becomes the multi-fractal spectrum, and this process is also the process of finding the distribution of $\alpha$. This function describes the global regularity of the structure $S$. The multi-fractal spectrum can be assumed to be the fractal dimension on the subset characterized by $\alpha$\cite{fenxing4}:
\begin{equation}\label{equ7}
f_{\varepsilon}\left(\alpha_i\right)=-\frac{\ln \left(N_{\varepsilon}\left(\alpha_i\right)\right)}{\ln (\varepsilon)}.
\end{equation}
$N_{\varepsilon}\left(\alpha_i\right)$ is the number of boxes $S_i$ containing a particular value $\alpha$. The limit can be obtained from the above equation\cite{fenxing4}:
\begin{equation}\label{equ8}
f(\alpha)=\lim _{\varepsilon \rightarrow 0}\left(f_{\varepsilon}(\alpha)\right).
\end{equation}

The multi-fractal spectrum $f(\alpha)$ computed by the above procedure becomes the Hausdorff dimension\cite{hossdoff1} of the $\alpha$ distribution. The multi-fractal spectrum describes the local and global patterns of the structure under study. Therefore, multi-fractal analysis can be used to characterize and extract certain features hidden in large amounts of data.



\subsection{Fractal Analysis in Super-resolution}

In this study, we accurately calculate the fractal dimensions of different high-resolution images and their corresponding low-resolution images for some images of the FFHQ\cite{FFHQ}, CelebA\cite{celeba}, and DIV2K\cite{df2k} datasets. We found that the fractal dimension of the high-resolution images is generally higher than that of the low-resolution images. This is shown in table \ref{tab:0}. This phenomenon indicates that there is a difference in fractal dimension between high resolution images and low resolution images. It can to some extent reflect the characteristics of the image in terms of detail richness, texture complexity, and so on.

\begin{table}[h]
  \caption{The fractal dimensions of the high-resolution image and the low-resolution image were calculated for 100 images randomly selected from each of the FFHQ, CelebA, and DIV2K datasets. FD denotes the fractal dimension and DIFF denotes the difference between the fractal dimension of the high and low resolution images.}
  \centering
  \begin{tabularx}{0.85\linewidth}{>{\raggedright\arraybackslash}p{2cm}*{3}{>{\centering\arraybackslash}X}} 
    \toprule
     Dataset& FD of HR & FD of LR & DIFF \\
    \midrule
    FFHQ &  2.4093& 2.4674 &  0.0581\\
    CelebA & 2.3346&  2.2644&  0.0702\\
    DIV2K &  2.7012&  2.6181&  0.0831 \\
    \bottomrule
  \end{tabularx}
  \label{tab:0}
\end{table}

In multi-fractal analysis, the fractal dimension at different $\alpha$ provides detailed information about the singular behavior of the system at different intensities or scales. Multi-fractal analysis involves dividing the space into multiple point sets $E_{\alpha}$ based on some categorical term $\alpha$. For each point set $E_{\alpha}$, i.e., the set of all points with the same $\alpha$, let $dim(E_{\alpha})$ denote its fractal dimension. Multi-fractal spectrum is given by the multi-fractal function $dim(E_{\alpha})$ vs $\alpha$.

In general, $\alpha$ is known as the singularity index, and the fractal dimension $f(\alpha)$ denotes the fractal dimension of a subset with the same degree of singularity. All $\alpha$ corresponding fractal dimensions $f(\alpha)$ constitute the multi-fractal spectrum. The multi-fractal spectrum can comprehensively characterize the internal structural heterogeneity of a complex system. By analyzing the shapes (e.g., widths, peaks, etc.) of the multi-fractal spectrum, the distribution of complexity at different levels of the system can be understood.

\begin{figure}[h]
    \centering
    \includegraphics[width=0.5\textwidth]{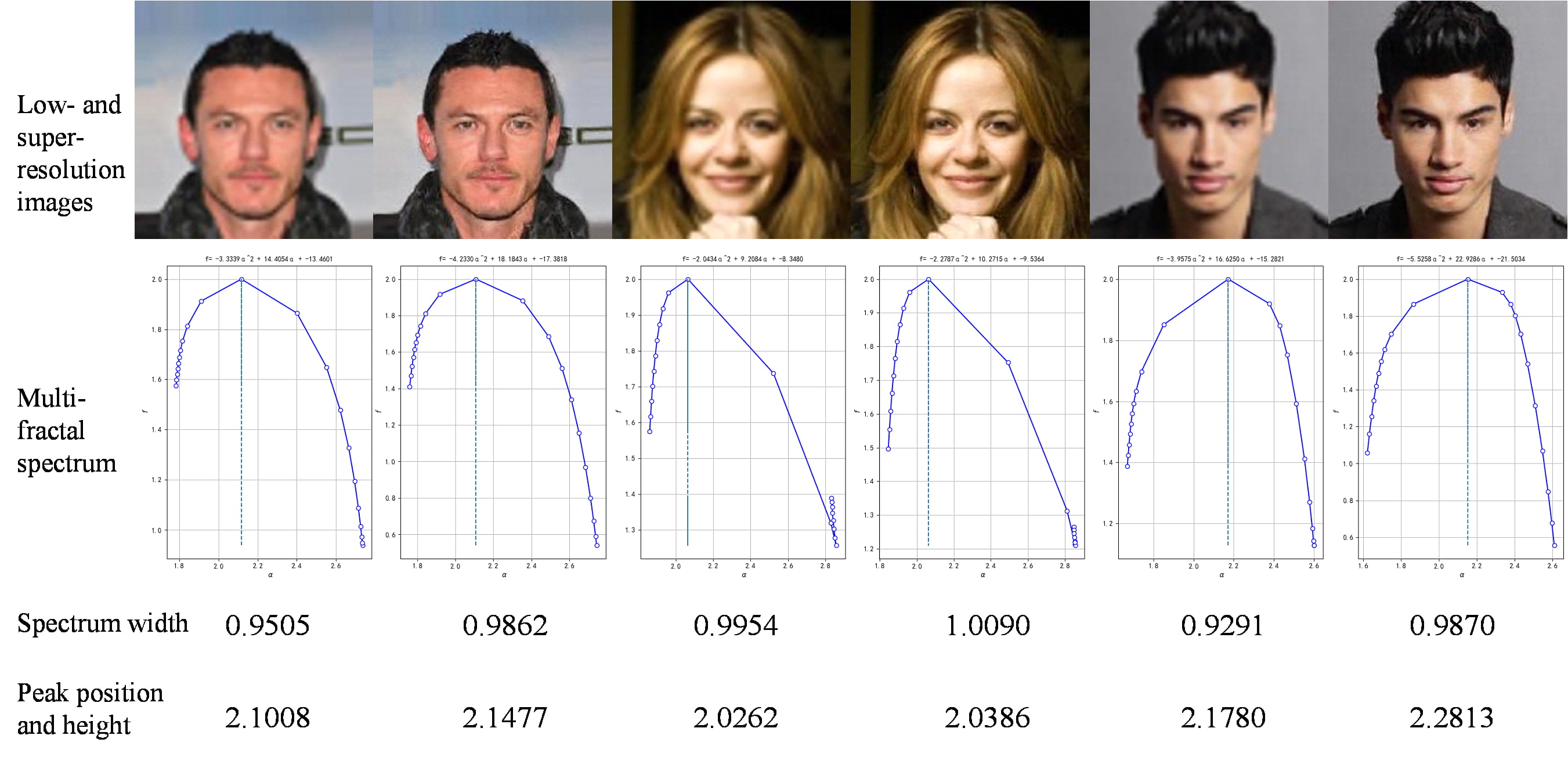}
    \caption{This figure shows 3 low-resolution images and the corresponding super-resolution images. The corresponding multi-fractal spectrum images are shown below the corresponding images and annotated with the corresponding fitted quadratic functions.}
    \label{fig:10}
\end{figure}

Figure \ref{fig:10} shows three low-resolution and super-resolution images and the corresponding multi-fractal spectra, and presents some key data. Higher resolution images contain more details and hence the local complexity of the image varies over a wider range and hence the multi-fractal spectrum will be wider. The richly detailed texture causes the peak position of the multi-fractal spectrum to shift in the direction of higher values. Thus the high singularity portion of the high resolution image is more prominent. It can be seen that there is a big difference between low resolution images and high resolution images in terms of multi-fractal, so it is reasonable to enhance the super-resolution effect from the point of view of fractal features.

\subsection{Diffusion Model}

In this section, we briefly review diffusion modeling.

DDPM\cite{DDPM1} is a method for modeling data distribution using the diffusion process. The diffusion process consists of two phases: a forward diffusion phase and a backward diffusion phase. In the training phase, the image $x_0$ is transformed into Gaussian noise by gradually adding $T$-step noise to the original image. The process of adding noise at each step can be represented as:

\begin{equation}\label{equ1}
q\left(x_t \mid x_{t-1}\right)=\mathcal{N}\left(x_t ; \sqrt{1-\beta_t} x_{t-1}, \beta_t I\right),
\end{equation}
where $x_t$ represents the noise image at time $t$, and $\beta_t$ is a hyperparameter. Through the reparameterization technique, $x_t$ can be directly obtained by $x_0$ :
\begin{equation}\label{equ2}
\quad q\left(x_t \mid x_0\right)=\mathcal{N}\left(x_t ; \sqrt{\overline{\alpha}_t} x_0,\left(1-\overline{\alpha}_t\right) I\right),
\end{equation}
where $\alpha_t=1-\beta_t$ and $\overline{\alpha}_t=\prod_{i=1}^t \alpha_i$. In the inference stage, the diffusion model will sample the Gaussian noise $x_T$, and $x_T$ is obtained by the denoising device to obtain $x_{T-1}$, and so on, gradually denoising until a high-quality output $x_0$ is obtained :
\begin{equation}\label{equ3}
p_\theta\left(x_{t-1} \mid x_t\right)=\mathcal{N}\left(x_{t-1} ; \mu_\theta\left(x_t, t\right), \Sigma_\theta\left(x_t, t\right)\right),
\end{equation}
where mean $\mu_t\left(\mathbf{x}_t, \mathbf{x}_0\right)=\frac{1}{\sqrt{\alpha_t}}\left(\mathbf{x}_t-\epsilon \frac{1-\alpha_t}{\sqrt{1-\bar{\alpha}_t}}\right)$ and variance $\sigma_t^2=\frac{1-\bar{\alpha}_{t-1}}{1-\bar{\alpha}_t} \beta_t$. Therefore, the noise image in step $t$ can be denoised to generate the noise image in step $t-1$ until the original image $x_0$ is obtained. In the reverse process, U-Net\cite{unet} is usually selected as the denoiser.

In order to be able to train the denoiser $\epsilon_{\theta}(x_t,t)$, for a given clean image $x_0$, the DDPM randomly samples a time step and a noise. It generates an image containing the noise according to Eq.\ref{equ2}. Then the model optimizes the parameters of the denoiser $\epsilon_{\theta}$:
\begin{equation}\label{equ4}
\nabla_{\boldsymbol{\theta}}\left\|\epsilon-\epsilon_{\boldsymbol{\theta}}\left(\sqrt{\bar{\alpha}_t} x_0+\epsilon \sqrt{1-\bar{\alpha}_t}, t\right)\right\|_2^2.
\end{equation} 

\begin{figure*}[h]
    \centering
    \includegraphics[width=\textwidth]{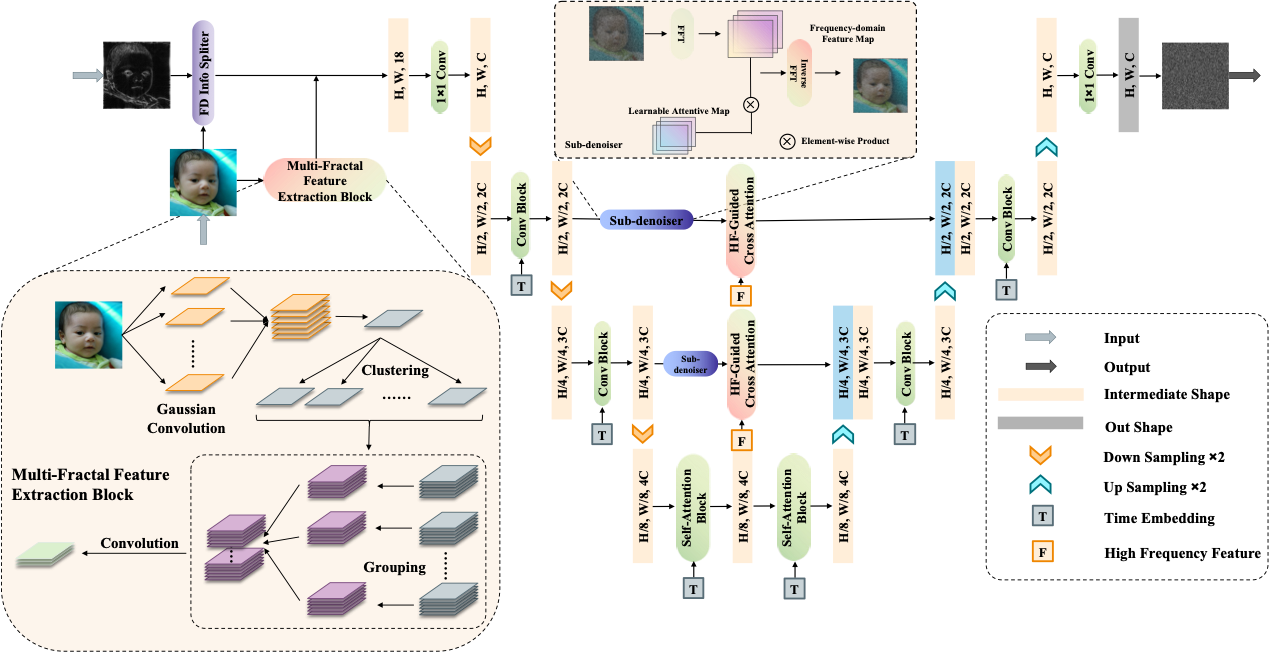}
    \caption{Diagram of the denoiser structure in MFSR. This figure highlights the overall U-Net structure, the multi-fractal analysis module and the sub-denoiser locations. FD Info Spliter reference from \cite{resdiff}.}
    \label{fig:2}
\end{figure*}

\section{Proposed Method}

Since the computation of fractal features involves traversing the image with windows of varying scales, the process is analogous to convolution. Therefore, we use convolution to approximate the original acquisition of fractal features. These acquired features serve as additional conditional inputs to the diffusion model's denoiser, enhancing the model's ability to recover detailed textures. Additionally, an attention-based sub-denoiser is introduced to further enhance the model's generative capability.

\subsection{MFSR}

The multifractal features of the image that are approximated using convolution are used as additional a priori conditions in the super-resolution process. For the denoiser, we select a U-Net structure based on the residual architecture with a Transformer, which has demonstrated strong denoising capabilities in numerous experiments. Additionally, we introduce a sub-denoiser within the U-Net cascade. The denoiser structure is illustrated in Figure \ref{fig:2}. The next section details the specific implementation method.

\subsection{MF Feature Extraction Block}

This section focuses on methods that utilize similar multi-fractal spectral feature extraction of images. Multi-fractal spectral analysis is a well-validated mathematical tool that captures statistical invariant features\cite{bubianxing} following a power law criterion to encode image self-similarity. It describes the global properties of self-similar objects that adhere to the power law criterion as the scale changes, reflecting the global spatial distribution characteristics of the image. Based on the approach of multi-fractal analysis in texture recognition\cite{encoding}, we propose a Multi-fractal Feature Extraction Block (MFB) to generate rich self-similarity information in space.

\begin{figure}[h]
    \centering
    \includegraphics[width=0.4\textwidth]{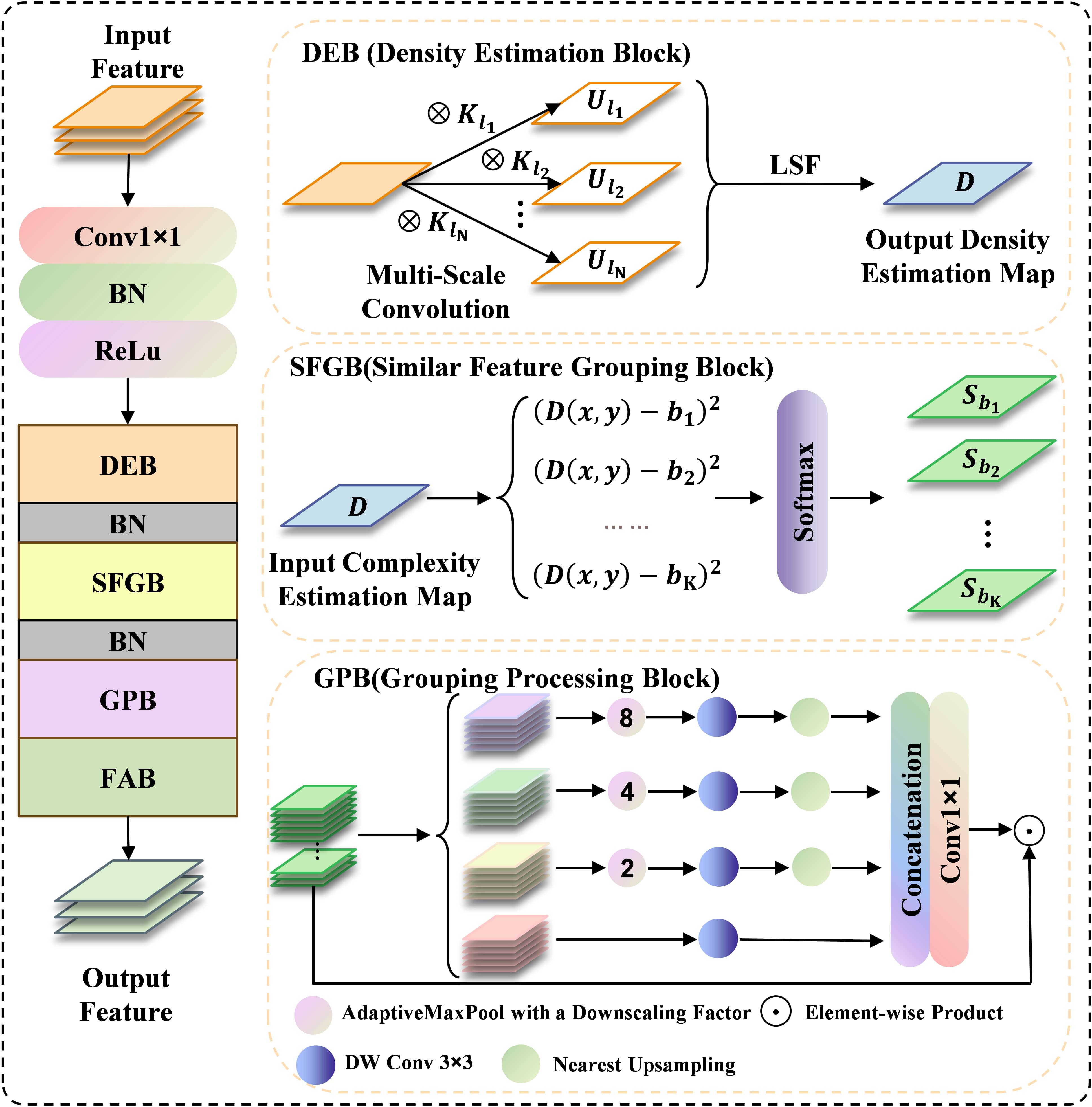}
    \caption{The left side represents the MFB structure. The input feature map first passes through the 1$\times$1 convolutional layer, BN layer, and ReLU. And then passes through the CEB, SFGB, GPB, and FAB to finally obtain the output feature map. The right side shows the detailed structure of DEB, SFGB and GPB respectively.}
    \label{fig:3}
\end{figure}

Fractal features capture the self-similarity and complexity of an image, providing a comprehensive description that includes information about the texture, shape, and structure of the image. When computing the multi-fractal spectrum, it is necessary to create an image computational density map and group the image pixels based on this map. This process effectively groups features of different textures. Therefore, incorporating multi-fractal spectrum computation into a super-resolution generative model can enhance the processing of detailed textures. However, implementing the image density map computation involves a nested for loop to traverse each pixel and calculate the density of neighboring regions. Directly adding this process to the model would significantly slow down the neural network's training. To address this, we approximate part of the multi-fractal spectrum computation using a method more suitable for neural network models, as illustrated in Figure \ref{fig:3}.

\begin{figure}[h]
    \centering
    \includegraphics[width=0.5\textwidth]{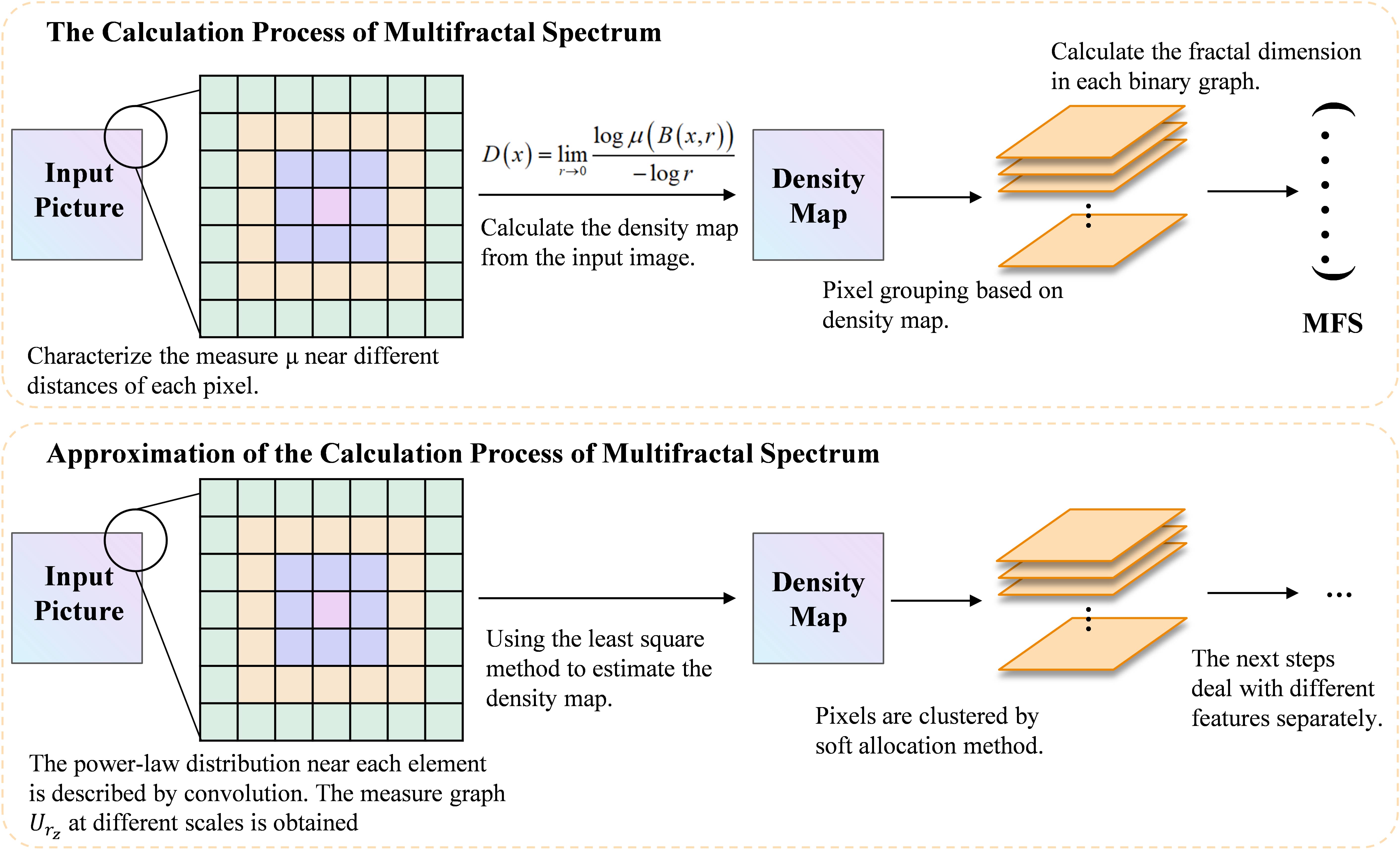}
    \caption{This figure shows a comparison of the multi-fractal calculation process with the one approximated in this paper.}
    \label{fig:4}
\end{figure}

In the computation of the multi-fractal features, it is necessary to define the statistic $\mu$ that obeys a power-law distribution along the scale direction. $D(x)$ inscribes the local power-law distribution in the neighbourhood of $x$ in terms of the measure $\mu$:
\begin{equation}\label{equ9}
D(x)=\lim _{r \rightarrow 0} \frac{\log \mu(B(x, r))}{-\log r},
\end{equation}
where $x \in I$ and $I$ is the input image. $B(x, r)$ denotes the hypersphere centred at $x$ with radius $r$, and $D(x)$ denotes the density in the vicinity of $x$. In the Figure \ref{fig:4}, during the computation of the multi-fractal spectrum, it is necessary to measure the power law distribution of regions of unused size centred on $x$ with different $r$ radii. This process is similar to the process of convolution. Specifically speaking, $x$ is the center of convolution, and different sizes of convolution kernels are chosen to approximate the power law distribution of different region sizes in the process of multi-fractal computation. Equivalently, the density map $D(x)$ can be obtained. After obtaining the density map $D(x)$, pixel points obeying the same power law distribution are clustered. Thus each sub-set of points has similar features, which in turn calculates the fractal dimensions of each sub-set of points in series as a vector of fractal dimensions, i.e., the multi-fractal spectrum. 

Approximatively, we use clustering to cluster the elements in the density map $D(x)$ into different sub-feature maps. We extract features for different feature subsets separately, thereby fusing the texture features of the image into the super-resolution method.

We call the previously described module for obtaining image texture features as MF Feature Extraction Block (MFB). In simple terms, MFB introduces the approximate multi-fractal information of the image into the denoising network, divides the feature map by point clustering, groups the divided features, and finally aggregates the obtained feature map into the denoising network. Therefore, MFB contains four main steps:


\begin{itemize}

\item Density Estimation Block (DEB): Follow the power law criterion to describe the local features of the image. In this part, this local feature is expressed as local density.

\item Similar Feature Group Block (SFGB): Divide the points on the feature map into different point sets of sub-feature maps according to the complexity.

\item Group Processing Block (GPB): All point set sub-feature maps are grouped and features are extracted by grouping.

\item Feature Aggregation Block (FAB): Aggregate the point set features obtained from group processing and connect them with the backbone network.
\end{itemize}

These four modules will be explained in detail in the following sections of this part.

\textbf{Density Estimation Block (DEB)}   The structure of Density Estimation Block is shown in Figure\ref{fig:3}. The input of this block is an image that has undergone simple low-frequency and high-frequency information recovery by Simple CNN, and the density estimation feature map will be obtained after this block.

The estimation of the density map defined by the estimation of Eq.\ref{equ9} is numerically realised by a least-squares fit of the metric map $Ur$ as well as the radii in a logarithmic coordinate system. By selecting different radii $r_1,r_2,... ,r_z$, a series of metric maps $Ur$ can be obtained.According to \cite{chongfenxing1}, $B(x,r)$ is a hypersphere centred at $x$ with radius $r$, and $\mu$ is the metric function. The metric function at the centre of $x$ can be described as:
\begin{equation}\label{equ11}
\mu(x, r)=k r^{d(x)}.
\end{equation}
Taking the logarithm of both sides of the equation, the density estimation problem can be transformed into a least squares problem, and according to Eq.\ref{equ9}, we can get:
\begin{equation}\label{equ14}
\min _{\boldsymbol{D}, \boldsymbol{K}} \sum_{r=1}^R\left(\boldsymbol{D}(u, v) \log l_r-\log \boldsymbol{U}_{l_r}(u, v) + \boldsymbol{K}(u, v) \right)^2,
\end{equation}
where $\boldsymbol{K}$ is the bias term. The measure $\mu$ calculates a value for each local block $I_i$, and for each channel of the feature map. $I_i(x,r)$ can be regarded as a local block of size $r \times r$ in the feature map. In the process of fractal dimension calculation, different local blocks in the image need to be processed one by one. The process is very similar to the convolution process of neural networks and hence convolution can be utilised to the process instead \cite{encoding}. In turn, the metric map is constructed.
The density map $\boldsymbol{D}$ can be estimated by taking the partial derivatives of $\boldsymbol{D}$ and $\boldsymbol{K}$, respectively. Let $\mathbf{a}$ be the vector whose $r$-th element is $a_r=\log l_r$ and $\mathbf{b}$ be the vector whose $r$-th element is $b_r=\log \boldsymbol{U}_{l_r}(u, v)$. Hence, for different radii $l_1,l_2,... ,l_R$, a series of metric maps $\boldsymbol{U}r$ can be obtained. 

The density map $\boldsymbol{D}$ of image $x$ at $(u, v)$ can be expressed as:

\begin{equation}\label{equ22_vector}
\boldsymbol{D}(u, v) = \frac{R\mathbf{a}^T\mathbf{b}-sum(\mathbf{a}\mathbf{b}^T)}{R\mathbf{a}^T\mathbf{a}+sum(\mathbf{a}\mathbf{a}^T)},
\end{equation}
where $sum(\cdot)$ represents summing up all the elements of the matrix, $\mathbf{a} = (a_1, a_2, \ldots, a_R)^T$ and $\mathbf{b} = (b_1, b_2, \ldots, b_R)^T$. The exact calculation procedure can be found in the Appendix A.






The structure is shown in Figure \ref{fig:3}.

\textbf{Similar Feature Grouping Block (SFGB)}   SFGB performs pixel point clustering based on the density map output from the density estimation module, and divides the original feature map into a set of feature map slices.

SFGB classifies the points on the output feature map $D$ obtained based on DEB for the purpose of similar feature clustering. Assuming that the size of the feature map $D$ is $M \times N$, the set of $ S = \{ 1, \dots, M \} \times \{ 1, \dots, N \} \subset R^2$. For any $k=1,2,\dots ,K$, for a given corridor $[C_1,C_2],\dots ,[C_K,C_{K+1}]$ obtain different subsets $S_{{C}_k}=\left\{(h, w) \in S|{C}_k \leq {D}(h, w)<{C}_{k+1}\right\}$. Each subset $S_{{C}_k}$ denotes whether each point in the set $S$ exists in the interval $[C_k,C_{k+1}]$ or not. If $C_k \leq D(u, v) \leq C_{k+1}$, then $D(u, v)$ = 1, otherwise $D(u, v)$ = 0. Thus, the computed $S_{{D}_k}$ is a bipartite map.

When performing the interval division work, the way the intervals are selected has a great influence on the results of the subsequent analysis\cite{influnece}. Since the graph $D$ generated in the previous stage has an uncertain range of values, different intervals will give different results. In addition, if the intervals are specified directly, not only the training error may increase, but also it is difficult to obtain the optimal intervals directly\cite{error}.

In view of this, we set the endpoints of the intervals as updatable parameters and use the soft assignment method to divide the intervals\cite{encoding} \cite{error}. This approach not only solves the problem of choosing the interval division scheme, but also benefits the optimization of backpropagation model training. this soft assignment approach uses probability values to represent the affiliation of $S_{b_k}(u, v)$ to an interval. Let $b_k(k = 1,\dots , K)$ denote a set of trainable interval anchors. Each $b_k$ has its corresponding affiliation mapping $S_{b_k} \in [ 0, 1 ] ^{M \times N}$ :


\begin{equation}\label{equ23}
\boldsymbol{S}_{b_k}(u, v)=\frac{\exp \left(-a_k\left(\boldsymbol{D}(u, v)-b_k\right)^2\right)}{\sum_{k=1}^K \exp \left(-a_k\left(\boldsymbol{D}(u, v)-b_k\right)^2\right)},
\end{equation}
where $a_k$ is a learnable parameter. This process can be realized by softmax. Thus, taking the feature map $D$ generated by DEB as input, a series of soft affiliation maps can be generated: $\left\{\boldsymbol{S}_{b_k} \in[0,1]^{M \times N}\right\}_{k=1}^K$. Intuitively, this hierarchical approach represents a grouping of different types of features. The structure is shown in Figure \ref{fig:3}.

\textbf{Grouped Processing Block (GPB)}  The SFGB section has been clustered for information of different densities. In the process of conditional image based image generation, regions with different densities are processed in different ways. Therefore, this part groups the clustering graphs obtained from the Similar Feature Clustering module for processing \cite{safm}. The structure is shown in Figure \ref{fig:3}.

First, depending on the feature grouping, the normalized input features are subjected to a channel splitting operation to produce a four-part component. Except for the first part of the component which is not processed, the remaining three parts of the component are input to the multiscale feature generation unit for processing. After processing, each part undergoes a DW convolution with a convolution kernel size of $3 \times 3$. The three components of the multiscale feature generation performed in the first step will be up-sampled to the original resolution by nearest interpolation of the resolution-specific features. Finally these multiscale features are then concatenated to aggregate local and global relationships by $1 \times 1$ convolution. Finally, we normalize them by GELU nonlinearity to estimate the attention map and adaptively adjust the complexity feature map based on the estimated attention.

\textbf{Feature Aggregation Block (FAB)}  This module aggregates the information obtained in the grouped feature processing module and inputs the obtained features into a conditional denoising network. Briefly, this part aggregates the grouped information through only one $3 \times 3$ convolution module.

\subsection{Attention-based noise reduction Block}  The inference process of the probabilistic diffusion model is to obtain a clean target image from a noisy map obeying a standard normal distribution by continuous denoising through a conditional denoiser. Thus for the conditional denoising network, the modified U-Net, there is a lot of portion of noise in the input. Fourier transform is a signal processing technique used to transform an image into the frequency domain. Through filtering operations, high frequency noise can be selectively removed to improve the image quality \cite{fft}. Therefore, we use a sub-denoiser based on combining attention with Fourier transform to minimize the effect of noise on the attention part \cite{medsegdiff}.

\begin{figure}[h]
    \centering
    \includegraphics[width=0.4\textwidth]{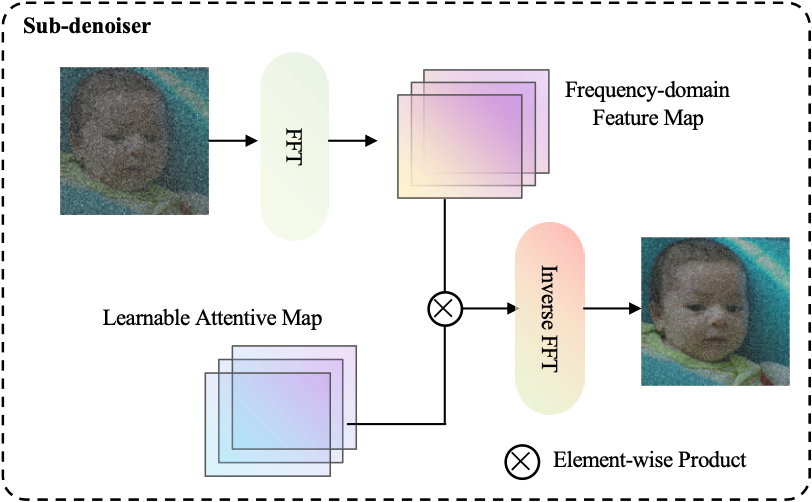}
    \caption{This is the schematic diagram of the sub-denoiser structure. First the feature map is transformed by FFT to get the frequency domain features. After element-by-element multiplication with a learnable attention map it is transformed to the spatial domain by IFFT. In this way, the feature map with high frequency noise removed is obtained.}
    \label{fig:5}
\end{figure}

The main role of the sub-denoiser is to constrain the noise component of the downsampled portion of the U-Net, and the structure is shown in Figure \ref{fig:5}.

\begin{table*}[h]
  \caption{Results of qualitative comparisons based on 4$\times$ super-resolution for the FFHQ dataset and 8$\times$ super-resolution for the CelebA dataset. Where bold values indicate the best values for each column of metrics and underlined values represent the suboptimal values for each column of metrics. For some of the experimental data we reused the results from \cite{resdiff}.}
  \centering
  \setlength{\tabcolsep}{3pt} 
  \begin{tabularx}{1.0\linewidth}{>{\raggedright\arraybackslash}p{2cm}*{12}{>{\centering\arraybackslash}X}} 
    \toprule
    & \multicolumn{6}{c}{FFHQ 4 $\times$} & \multicolumn{6}{c}{CelebA 8 $\times$} \\
    \cmidrule(lr){2 - 7} \cmidrule(lr){8 - 13}
    & PSNR$\uparrow$ & SSIM$\uparrow$ & FID$\downarrow$ &LPIPS$\downarrow$ & CLIPIQA$\uparrow$ & MUSIQ$\uparrow$ & PSNR$\uparrow$ & SSIM$\uparrow$ & FID$\downarrow$ &LPIPS$\downarrow$ &CLIPIQA$\uparrow$& MUSIQ$\uparrow$ \\
    \midrule
    Ground Truth & $\infty$ & 1.000 & 0.00 & 0.00 & 1.000 & $\infty$ & $\infty$ & 1.000 & 0.00 & 0.00 & 1.000 & $\infty$\\
    \midrule
    SRGAN\cite{srgan1} & 17.57 & 0.688 & 156.07 & 0.475 & 0.4773 & 44.133 & - & - & - & - & - & - \\
    ESRGAN\cite{esrgan1} & 15.43 & 0.267 & 166.36 & 0.367 & 0.4525 & 42.151  & 23.24 & 0.66 & 84.13 & 0.3185 & 0.4871 & 44.847 \\
    BRGM\cite{brgm} & 24.16 & 0.70 & 81.38 & 0.257 & 0.5137 & 47.615 & - & - & - & - & - & - \\
    PULSE\cite{pulse}& 15.74 & 0.37 & 168.94 & 0.461 & 0.4533 & 44.094 & - & - & - & - & - & - \\
    SRDiff\cite{srdiff} & 26.07 & 0.794 & 72.36 & 0.2516 & 0.5846 & 50.106 & 25.31 & 0.73 & 80.98 & 0.2591 & 0.5413 & 48.082 \\
    SR3\cite{sr3} & 25.37 & 0.778 & 75.29 & 0.2583 & 0.5385 & 49.03 & 24.89 & 0.728 & 83.11 & 0.2683 & 0.5209 & 46.9602 \\
    PASD\cite{pasd} & \textbf{27.32} & 0.7830 & \uline{50.53} & 0.3517 & \uline{0.6208} & \uline{66.538} & 22.76 & 0.639 & \textbf{60.79} & 0.4216 & \uline{0.628} & \uline{58.573} \\
    ResShift\cite{resshift} & 26.64 & 0.781 & \textbf{50.08} & 0.4409 & 0.5859 & 63.109 & - & - & - & - & - & - \\
    DiffBIR\cite{diffbir} & 26.83 & 0.796 & 54.53 & 0.4097 & \textbf{0.7102} & \textbf{70.472} & 22.31 & 0.686 & \uline{62.53} & 0.4107 & \textbf{0.792} & \textbf{68.581} \\
    ResDiff\cite{resdiff} & 26.73 & \uline{0.818} & 70.54 & \uline{0.2486} & 0.5893 & 50.749 & \uline{25.37} & \uline{0.734} & 78.52 & \textbf{0.2547} & 0.5366 & 48.467 \\
    MFSR & \uline{26.94} & \textbf{0.833} & 69.64 & \textbf{0.2471} & 0.5976  & 51.157 & \textbf{25.39} & \textbf{0.735} & 77.94 & \uline{0.2579} & 0.5574 & 48.215\\
    \bottomrule
  \end{tabularx}
  \label{tab:1}
\end{table*}

\begin{figure*}[t]
    \centering
    \includegraphics[width=1\textwidth]{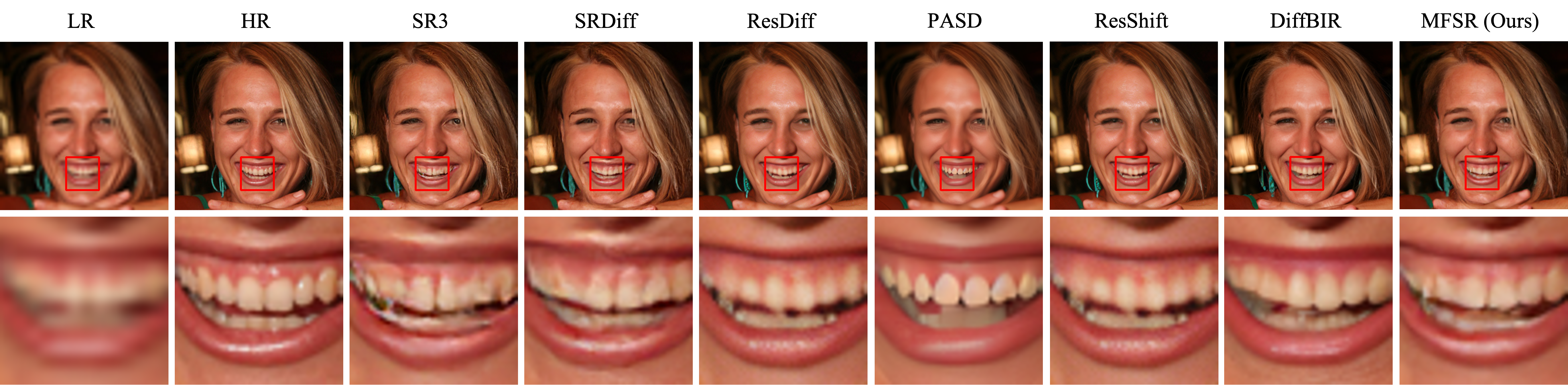}
    \caption{Comparisons of the 4× super-resolution based on FFHQ among SR3, SRDiff, PASD, ResShift, DiffBIR, ResDiff, and MFSR. It also shows the local zoom effect.}
    \label{fig:6}
\end{figure*}

The main role of the sub-denoiser is to constrain the noise component in the downsampled part of the U-Net, and the structure is shown in Figure \ref{fig:5}. The implementation is as follows: the feature map is subjected to a two-dimensional Fast Fourier Transform (FFT) in the spatial dimension, then a parametric attention mechanism applied to the features of the Fourier space is learned. The noise is reduced by multiplying the parametric mapping by the features of the Fourier space, and finally transferred back to the spatial domain by the Inverse Fast Fourier Transform(Inverse FFT). As shown in Eq.\ref{equ27}. $m$ is the feature map, $\mathcal{F}$ and $\mathcal{F}^{-1}$ denote the FFT and the Inverse FFT, respectively, A denotes the learnable attention mechanism, and $m$ denotes the feature map that has been inverted by the FFT.

\begin{equation}\label{equ27}
m^{\prime}=\mathcal{F}^{-1}\left[A \otimes \mathcal{F}[m]\right]
\end{equation}

Unlike conventional frequency filters, the attention mechanism is a learnable version and can be tuned globally for frequency components. So that it can learn to globally tune specific frequency components, constrain high-frequency components, and perform adaptive integration.

\section{Experiments}

In this section, we present the details of the experiment in terms of Dataset, Configuration, Performance and Ablation Study and analyze the results.

\subsection{Datasets}

MFSR performs different configurations of super-resolution experiments on face image data versus natural image data.

For face image super-resolution, we chose FFHQ\cite{FFHQ} with CelebA\cite{celeba} dataset. CelebA (Celeb Faces Attributes Dataset) dataset is a large-scale dataset containing more than 200k face images, and FFHQ (Flickr Faces Hight Quality) dataset is a dataset containing 70k face images. We directly downsampled the HR images with bicubic kernels and selected 5000 images as the test set and all other images as the training set.

For natural image super-resolution, we use the DIV2K\cite{df2k} and Urban100\cite{urban}. They are composed of content-rich natural images. In the training phase, we crop the HR images into 160 $\times$ 160 image chunks and downsample them using bicubic kernels. In the testing phase, we downsample the original HR images to obtain SR prediction images based on LR images. The DIV2K test set uses its given test set of 100 images, while for Urban100, we select 20 images as the test set.

\subsection{Configure}

The initial channels of U-Net are set to 64, and the channel number expansion multiplicity is 1, 2, 4, 8, 8. The self-attention module is only added to the bottom and penultimate layers, and each convolution block in the structure of U-Net is replaced by using two residual convolution blocks, and the dropout rate is set to 0.2. In the process of image fractal analysis, we set the number of clustering kernels in the similar feature clustering module to 64 and the final number of channels obtained from the convolution is 3. Subsequent ablation experiments have partially demonstrated that our hyperparameter settings are optimal.

During training, we used the Adam\cite{adam} optimizer with the learning rate set to $1 \times 10^{-4}$ and the batch size set to 4. The total time step $T$ was set to 1000, and $\beta_{1 : T}$ was increased linearly from $1 \times 10 ^ {- 6}$ to $1 \times 10 ^ {- 2}$. Training was performed on two NVIDIA 4090s.




\subsection{Performance}

\begin{table*}[h]
  \caption{Results of qualitative comparisons based on 4$\times$ super-resolution for the Urban100 dataset. Where bold values indicate the best values for each column of metrics and underlined values represent the suboptimal values for each column of metrics. For some of the experimental data we reused the results from \cite{resdiff}.}
  \centering
  \setlength{\tabcolsep}{3pt} 
  \begin{tabularx}{1.0\linewidth}{>{\raggedright\arraybackslash}p{2cm}*{12}{>{\centering\arraybackslash}X}} 
    \toprule
    & \multicolumn{6}{c}{DIV2K 4 $\times$} & \multicolumn{6}{c}{Urban100 4 $\times$} \\
    \cmidrule(lr){2 - 7} \cmidrule(lr){8 - 13}
    & PSNR$\uparrow$ & SSIM$\uparrow$ & FID$\downarrow$ &LPIPS$\downarrow$ & CLIPIQA$\uparrow$ & MUSIQ$\uparrow$ & PSNR$\uparrow$ & SSIM$\uparrow$ & FID$\downarrow$ &LPIPS$\downarrow$ &CLIPIQA$\uparrow$& MUSIQ$\uparrow$ \\
    \midrule
    Ground Truth & $\infty$ & 1.000 & 0.00 & 0.00 & 1.000 & $\infty$ & $\infty$ & 1.000 & 0.00 & 0.00 & 1.000 & $\infty$\\
    \midrule
    SRDiff\cite{srdiff} & 26.87 & 0.69 & 110.32 &  0.253 & 0.5617  & 53.117 & 26.49 & 0.79 & 51.37 & 0.257 & 0.568 & 51.604 \\
    SR3\cite{sr3} & 26.17 & 0.65 & 111.45 & 0.271 & 0.5476 & 52.063 & 25.18 & 0.62 & 61.14 & 0.263 & 0.533 & 48.932 \\
    PASD\cite{pasd} & 21.59 & 0.51 & \textbf{52.13} & 0.422 & \uline{0.6137} & \uline{62.879} & 20.02 & 0.47 & 63.42 & 0.428 & 0.529 & 50.632 \\
    ResShift\cite{resshift} & 24.16 & \textbf{0.74} & \uline{56.59} & 0.427 & 0.592 & 58.017 & 20.42 & 0.55 & 52.19 & 0.437 & 0.606 & 60.108 \\
    DiffBIR\cite{diffbir} & 23.68 & 0.72 & 59.69 & 0.397 & \textbf{0.718} & \textbf{72.401} & 21.24 & 0.65 & 57.58 & 0.412 & \textbf{0.705} & \textbf{69.581} \\
    ResDiff\cite{resdiff} & \uline{27.94} & 0.72 & 106.71 & \uline{0.246} & 0.5707 & 54.706 & \uline{27.43} & \uline{0.82} & \uline{42.35} & \uline{0.248} & \uline{0.583} & \uline{54.827}\\
    MFSR & \textbf{28.16} & \uline{0.73} & 106.79 & \textbf{0.233} & 0.5820 & 55.143 & \textbf{27.66} & \textbf{0.83} & \textbf{40.91} & \textbf{0.242} & 0.594 & 55.109\\
    \bottomrule
  \end{tabularx}
  \label{tab:2}
\end{table*}

\begin{figure*}[h]
    \centering
    \includegraphics[width=1\textwidth]{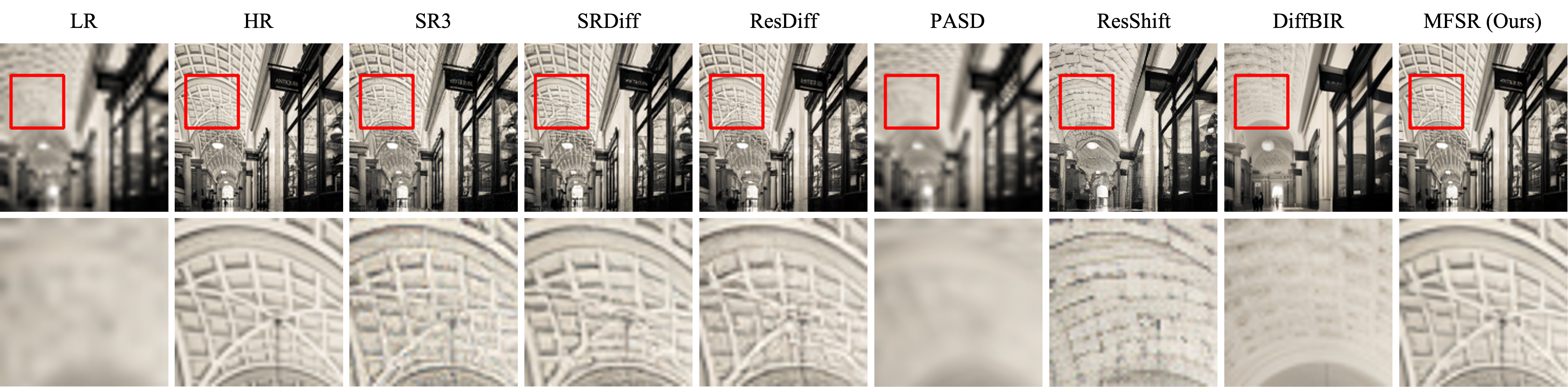}
    \caption{Comparisons of the 4$\times$ super-resolution based on DIV2K and Urban100 among SR3, SRDiff, PASD, ResShift, DiffBIR, ResDiff, and MFSR. It also shows the local zoom effect.}
    \label{fig:13}
\end{figure*}

\textbf{Face-based super-resolution}  Based on the face datasets FFHQ and CelebA, we validate on different 4$\times$ and 8$\times$ super-resolution specifications. The validation results of 4$\times$ super-resolution based on FFHQ and 8$\times$ super-resolution based on CelebA are shown in the Table \ref{tab:1}. The metrics selected for comparison are PSNR\cite{psnr}, SSIM\cite{psnr}, FID\cite{fid} and LPIPS\cite{LPIPS}. In addition, there are non-reference  indicators CLIPIQA\cite{CLIPIQA} and MUSIQ\cite{MUSIQ}. From the experimental results in the table, it can be seen that MFSR performs better than other methods in some of the metrics. Please refer to Figure \ref{fig:6} for the super-resolution effect and details.

\textbf{Super-resolution based on natural images}  Based on the natural image datasets DIV2K and Urban100, we validate on a 4$\times$ super-resolution specification. The results are shown in Table \ref{tab:2}. The experimental results show that some experimental results of MFSR outperform the other models. Please refer to Figure \ref{fig:13} for the super-resolution effect and details.



\subsection{Additional Experimental Results}
To verify the generalisability of the model, we performed super-resolution performance tests on other datasets. We use MFSR for testing on the Set5\cite{set5} and Set14\cite{set14} datasets. Set5 and Set14 have high-quality images covering different categories, which are used to evaluate the ability of super-resolution algorithms to recover details. The experimental results are shown in Table \ref{tab:exp_text_1}.

\begin{table}[h]
  \caption{Results of qualitative comparisons based on 4$\times$ super-resolution for the Set5 dataset and Set14 dataset. Where bold values indicate the best values for each column of metrics and underlined values represent the suboptimal values for each column of metrics.}
  \centering
  \begin{tabularx}{1.0\linewidth}{>{\raggedright\arraybackslash}p{2cm}*{4}{>{\centering\arraybackslash}X}} 
    \toprule
    & \multicolumn{2}{c}{Set5 4$\times$} & \multicolumn{2}{c}{Set14 4$\times$} \\
    \cmidrule(lr){2 - 3} \cmidrule(lr){4 - 5}
    & PSNR$\uparrow$ & SSIM$\uparrow$ & PSNR$\uparrow$ & SSIM$\uparrow$\\
    \midrule
    Ground Truth & $\infty$ & 1.000 & $\infty$ & 1.000 \\
    \midrule
    SRDiff\cite{srdiff} & 28.72 & 0.843 & 25.63 & 0.702 \\
    SR3\cite{sr3} & 27.31 & 0.767 & 25.29 & 0.684\\
    ResDiff\cite{resdiff} & \uline{29.32} & \uline{0.854} & \uline{26.19} & \uline{0.718} \\
    MFSR & \textbf{29.87} & \textbf{0.858} & \textbf{26.33} & \textbf{0.72} \\
    \bottomrule
  \end{tabularx}
  \label{tab:exp_text_1}
\end{table}

In addition, we performed the same comparison for each diffusion-based method on LSCIDMR\cite{LSCIDMR} and Manga109\cite{Manga109}. The LSCIDMR cloud image dataset contains a variety of cloud images, and Manga109 has 109 cartoon images. The two can be used to verify the super-resolution effect of the model on meteorological cloud images and cartoon images respectively. The results are shown in Table \ref{tab:exp_text_2}. The test results show that MFSR can outperform other models in most of the experimental results.

\begin{table}[h]
  \caption{Results of qualitative comparisons based on 4$\times$ super-resolution for the LSCIDMR dataset and Managa109 dataset. Where bold values indicate the best values for each column of metrics and underlined values represent the suboptimal values for each column of metrics.}
  \centering
  \begin{tabularx}{1.0\linewidth}{>{\raggedright\arraybackslash}p{2cm}*{4}{>{\centering\arraybackslash}X}} 
    \toprule
    & \multicolumn{2}{c}{LSCIDMR 4$\times$} & \multicolumn{2}{c}{Managa109 4$\times$} \\
    \cmidrule(lr){2 - 3} \cmidrule(lr){4 - 5}
    & PSNR$\uparrow$ & SSIM$\uparrow$ & PSNR$\uparrow$ & SSIM$\uparrow$\\
    \midrule
    Ground Truth & $\infty$ & 1.000 & $\infty$ & 1.000 \\
    \midrule
    SRDiff\cite{srdiff} & 27.54 & 0.807 & 27.04 & 0.813 \\
    SR3\cite{sr3} & 26.13 & 0.782 & 26.88 & 0.805\\
    ResDiff\cite{resdiff} & \uline{27.79} & \textbf{0.812} & \uline{27.76} & \uline{0.832} \\
    MFSR & \textbf{27.89} & \textbf{0.812} & \textbf{27.84} & \textbf{0.836} \\
    \bottomrule
  \end{tabularx}
  \label{tab:exp_text_2}
\end{table}

\subsection{Ablation Study}
In this part, we perform ablation experiments based on FFHQ32$\rightarrow$128. The first part examines the validity of each component through ablation experiments, and the second part examines the validity of hyperparameters through ablation experiments.

\textbf{Hyperparameter}  This part examines the influence of different hyperparameters on the experimental results, and selects different hyperparameters for the experiments. These include the number of clustering kernels in SFGB, the number of channels for MFB to generate feature maps, the number of denoising steps in the diffusion model and the number of channels in the denoising network. The experimental results are shown in the Table \ref{tab:4}. MFSR balances between efficiency and performance when the number of clustering kernels is chosen to be 64, the number of channels of MFB generating features is 3, the number of denoising steps is 1000, and the number of channels of denoising network is 64.

\begin{table}[h]
  \caption{Ablation experiments performed on FFHQ4$\times$ for different hyperparameters.}
  \centering
  \begin{tabularx}{\columnwidth}{>{\raggedright\arraybackslash}X@{\hspace{4pt}}>{\raggedright\arraybackslash}X@{\hspace{4pt}}>{\raggedright\arraybackslash}Xccc}
    \toprule
    \multicolumn{3}{c}{Hyperarameters} & \multicolumn{3}{c}{Merics} \\
    \cmidrule(lr){1 - 3} \cmidrule(lr){4 - 6}
    Clustering Kernels & Channels of MFB & Total Time Step & PSNR$\uparrow$ & SSIM$\uparrow$ & FID$\downarrow$ \\
    \midrule
    64 & 3 & 1000 & 26.94 & 0.833 & 69.64 \\
    \midrule
    16 & 3 & 1000 & 26.78 & 0.824 & 70.35 \\
    32 & 3 & 1000 & 26.83 & 0.827 & 70.17 \\
    128 & 3 & 1000 & 26.89 & 0.831 & 69.73 \\
    256 & 3 & 1000 & 26.76 & 0.819 & 69.86 \\
    \midrule
    64 & 6 & 1000 & 26.94 & 0.832 & 69.74 \\
    64 & 9 & 1000 & 26.94 & 0.831 & 69.81 \\
    64 & 12 & 1000 & 26.95 & 0.833 & 69.57 \\
    64 & 15 & 1000 & 26.95 & 0.831 & 69.60 \\
    \midrule
    64 & 3 & 50 & 24.53 & 0.742 & 76.39 \\
    64 & 3 & 200 & 25.67 & 0.791 & 74.63 \\
    64 & 3 & 2000 & 27.06 & 0.838 & 67.51 \\
    \bottomrule
  \end{tabularx}
  \label{tab:4}
\end{table}

\textbf{Framework}  This part examines the effectiveness of different components and ablation experiments are designed for different components, including MFB, DEB, GPB and Sub-denoiser. For the MFB module, we set up with and without MFB module to verify its effectiveness. In addition, to investigate the effectiveness of DEB in MFB, we used the methods with and without DEB for comparison. In order to examine the validity of the GPB module, we choose to perform the clustering process with and without the clustering method. In order to examine the effectiveness of sub-denoiser, we choose to compare the two methods using sub-denoiser with no sub-denoiser. The experimental results are shown in the Table \ref{tab:5}. The model has the best results in the case of choosing MFB, GPB, and SFGB, and the experimental results are worse than MFSR in the case of eliminating each part individually.

\begin{table}[h]
  \caption{Ablation experiments performed on FFHQ4$\times$ for different structures.}
  \centering
  \begin{tabularx}{\columnwidth}{Xccccccc}
    \toprule
    \multicolumn{4}{c}{Model   Components} & \multicolumn{3}{c}{Metrics} \\
    \cmidrule(lr){1 - 4} \cmidrule(lr){5 - 7}
    MFB & DEB & GPB & Sub-denoiser & PSNR$\uparrow$ & SSIM$\uparrow$ & FID$\downarrow$ \\
    \midrule
    $\checkmark$ & $\checkmark$ & $\checkmark$ & $\checkmark$ & 26.94 & 0.833 & 69.64 \\
    $\checkmark$ & $\checkmark$ & $\checkmark$ &  & 26.88 & 0.828 & 70.36 \\
    &  &  & $\checkmark$ & 26.79 & 0.820 & 70.31 \\
    & $\checkmark$ &  & $\checkmark$ & 26.79 & 0.82 & 70.22 \\
    &  & $\checkmark$ & $\checkmark$ & 26.80 & 0.815 & 70.47 \\
    \bottomrule
  \end{tabularx}
  \label{tab:5}
\end{table}

\begin{figure}[h]
    \centering
    \includegraphics[width=0.4\textwidth]{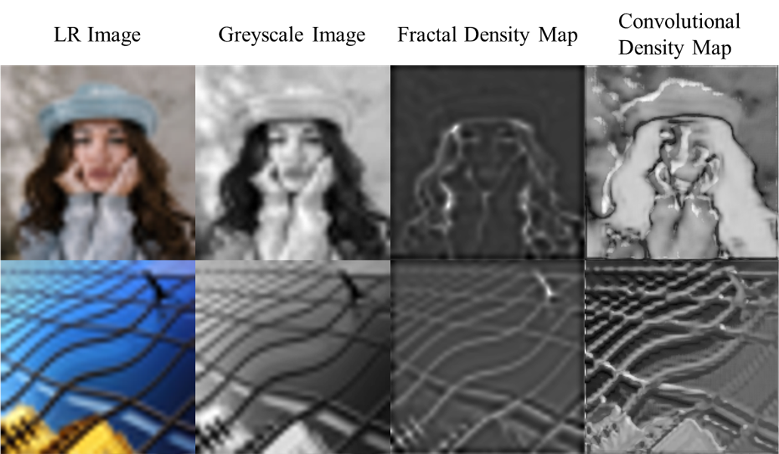}
    \caption{This figure presents a demonstration of the fractal density map. The first column is the original LR image, the second column is the corresponding grey scale image, the third column is the fractal density map obtained using the fractal dimension calculation method, and the last column is the fractal density map obtained by the convolution method approximation used in this paper.}
    \label{fig:8}
\end{figure}

\subsection{Visualisation}

Fractal features are capable of capturing rich information about micro and macro texture structures in images. In order to speed up the computational efficiency, this paper uses Multi-fractal Feature Extraction Block to approximate the fractal feature density map. Figure \ref{fig:8} visualises the original, grey scale, fractal density map, and convolutional approximation density map. It can be seen that the fractal density map mainly describes the edge texture information. Although the fractal density map approximated using the convolution method cannot be exactly the same as the fractal density map all the time, it can be achieved to obtain the approximate edge texture information.

\subsection{Texture Detail Optimization Analysis}
MFB, as an innovative design, endows the model with valuable a priori information about detailed textures. In order to rigorously verify this property, we select the widely recognized and representative DIV2K dataset for our experiments. Specifically, we comprehensively compare the MFSR model with and without MFB - MFSR$^{(1)}$. Meanwhile, to demonstrate the effectiveness of fractal computational methods for capturing texture features, we designed the texture prior acquisition model with only convolution - MFSR$^{(2)}$. The performance differences between the three models, texture prior with fractal texture prior, texture prior without fractal texture prior and texture prior with pure convolution, are analysed. The results are shown in Table \ref{tab:6}. Experimental results show that the approximate fractal feature prior has a proximate role in aiding the super-resolution effect, and that the convolution method alone does not outperform the fractal computational method.
\begin{table}[h]
  \caption{The table shows a comparison of the effects of MFSR and DIV2K 4$\times$ super-resolution using only the convolution method without the fractal method.}
  \centering
  \begin{tabular}{>{\centering\arraybackslash}p{0.3\linewidth}>{\centering\arraybackslash}c>{\centering\arraybackslash}c>{\centering\arraybackslash}c}
    \toprule
    \multirow{2}{*}{Model} & \multicolumn{3}{c}{Metrics} \\
    \cmidrule(lr){2 - 4}
      & PSNR$\uparrow$ & SSIM$\uparrow$ & FID$\downarrow$ \\
    \midrule
MFSR                  & \textbf{28.16}       & \textbf{0.73}       & \textbf{106.79}      \\
MFSR$^{(1)}$                 & 27.98       & 0.72       & 108.14      \\
MFSR$^{(2)}$                 & 28.02       & 0.73       & 109.70      \\
    \bottomrule
  \end{tabular}
  \label{tab:6}
\end{table}

\begin{figure}[h]
    \centering
    \includegraphics[width=0.3\textwidth]{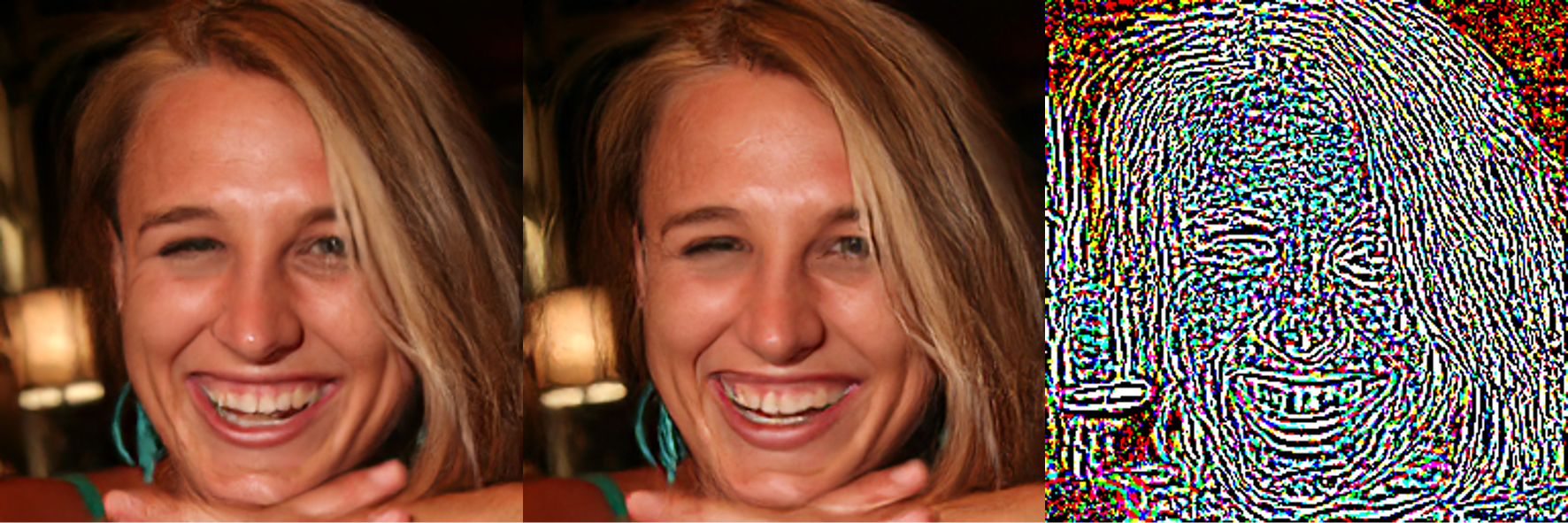}
    \caption{The first image shows the super-resolution results from MFSR. The second image shows the super-resolution results without MFB. The third image is the result of making a difference between the two.}
    \label{fig:14}
\end{figure}

Compared with the MFSR model without MFB integration, the MFSR model with MFB module shows superior performance. In order to visualise this difference, we have visualised the difference between the two super-resolution results, as shown in Fig\ref{fig:14}. From the visualisation results, it can be clearly observed that the main difference between the two models is concentrated in the edge texture region of the image. This phenomenon fully indicates that the re-fractal texture information extracted by the MFB module is of great significance in recovering the detailed texture of the image.

\section{Conclusion}
In this paper, we propose MFSR, a diffusion model-based image super-resolution method incorporating image multi-fractal features. The multi-fractal features of the image provide rich texture information for the reconstruction process of low-resolution images. Experiment results demonstrate that the inclusion of the multi-fractal features facilitates the super-resolution effect. Similarly, our extra texture information might be applied to other methods. For example, image classification, convolution-based super-resolution, and so on. Besides, the inference time of the model needs to be optimized continuously.


\begin{thebibliography}{1}
\bibliographystyle{IEEEtran}
\bibitem{super_resolution1}
D. C. Lepcha, B. Goyal, A. Dogra, et al., ``Image super-resolution: A comprehensive review, recent trends, challenges and applications,'' \textit{Information Fusion}, vol. 91, pp. 230--260, 2023. DOI: 10.1016/j.inffus.2022.10.007.

\bibitem{GAN1}
I. Goodfellow, J. Pouget-Abadie, M. Mirza, et al., ``Generative adversarial nets,'' \textit{Advances in neural information processing systems}, 2014. DOI: 10.1155/2019/2765120.

\bibitem{srgan1}
C. Ledig, L. Theis, F. Huszár, J. Caballero, A. Cunningham, A. Acosta, A. Aitken, A. Tejani, J. Totz, Z. Wang, W. Shi, ``Photo-Realistic Single Image Super-Resolution Using a Generative Adversarial Network,'' in \textit{2017 IEEE Conference on Computer Vision and Pattern Recognition (CVPR)}, 2017, pp. 105--114. DOI: 10.1109/CVPR.2017.19.

\bibitem{esrgan1}
X. Wang, K. Yu, S. Wu, et al., ``SRGAN: Enhanced Super-Resolution Generative Adversarial Networks,'' in \textit{ECCV 2018}, vol. 11133, pp. 63--79, 2018. DOI: 10.1007/978-3-030-11021-5\_5.

\bibitem{crash1}
C. Saharia, J. Ho, W. Chan, T. Salimans, D. J. Fleet, M. Norouzi, ``Image Super-Resolution via Iterative Refinement,'' \textit{IEEE Transactions on Pattern Analysis and Machine Intelligence}, vol. 45, no. 4, pp. 4713--4726, 2023. DOI: 10.1109/TPAMI.2022.3204461.

\bibitem{DDPM1}
J. Ho, A. Jain, P. Abbeel, ``Denoising diffusion probabilistic models,'' \textit{Advances in neural information processing systems}, vol. 33, pp. 6840--6851, 2020. DOI: 10.48550/arXiv.2006.11239.

\bibitem{hecheng1}
R. Rombach, A. Blattmann, D. Lorenz, et al., ``High-resolution image synthesis with latent diffusion models,'' \textit{Proceedings of the IEEE/CVF conference on computer vision and pattern recognition}, pp. 10684--10695, 2022.

\bibitem{hecheng2}
P. Dhariwal, A. Nichol, ``Diffusion models beat gans on image synthesis,'' \textit{Advances in neural information processing systems}, vol. 34, pp. 8780--8794, 2021.

\bibitem{hecheng3}
C. Saharia, W. Chan, S. Saxena, et al., ``Photorealistic text-to-image diffusion models with deep language understanding,'' \textit{Advances in Neural Information Processing System}, vol. 35, pp. 36479--36494, 2022.

\bibitem{hecheng4}
A. Ramesh, P. Dhariwal, A. Nichol, C. Chu, M. Chen, ``Hierarchical text-conditional image generation with clip latents,'' \textit{arXiv preprint arXiv:2204.06125}, vol. 1, no. 2, pp. 3, 2022.

\bibitem{xiufu1}
W. Liu, Y. Wang, K. -H. Yap, L. -P. Chau, ``Bitstream-Corrupted JPEG Images are Restorable: Two-stage Compensation and Alignment Framework for Image Restoration,'' in \textit{Proceedings of the IEEE/CVF Conference on Computer Vision and Pattern Recognition}, pp. 9979--9988, 2023.

\bibitem{xiufu2}
Y. Li, Y. Fan, X. Xiang, D. Demandolx, R. Ranjan, R. Timofte, L. Van Gool, ``Efficient and explicit modelling of image hierarchies for image restoration,'' in \textit{Proceedings of the IEEE/CVF Conference on Computer Vision and Pattern Recognition}, pp. 18278--18289, 2023.


\bibitem{sr3}
C. Saharia, J. Ho, W. Chan, T. Salimans, D. J. Fleet, M. Norouzi, ``Image super-resolution via iterative refinement,'' \textit{IEEE Transactions on Pattern Analysis and Machine Intelligence}, vol. 45, no. 4, pp. 4713--4726, 2022.

\bibitem{srdiff}
H. Li, Y. Yang, M. Chang, S. Chen, H. Feng, Z. Xu, Q. Li, Y. Chen, ``Srdiff: Single image super-resolution with diffusion probabilistic models,'' \textit{Neurocomputing}, vol. 479, pp. 47--59, 2022.

\bibitem{resdiff}
S. Shang, Z. Shan, G. Liu, J. Zhang, ``Resdiff: Combining cnn and diffusion model for image super-resolution,'' \textit{Proceedings of the AAAI Conference on Artificial Intelligence}, vol. 38(8), pp. 8975--8983, 2024.

\bibitem{unet}
O. Ronneberger, P. Fischer, T. Brox, ``U-net: Convolutional networks for biomedical image segmentation,'' in \textit{Medical Image Computing and Computer-Assisted Intervention--MICCAI 2015: 18th International Conference, Munich, Germany, October 5-9, 2015, Proceedings, Part III 18}, pp. 234--241, 2015.

\bibitem{flow}
D. P. Kingma, P. Dhariwal, ``Glow: Generative flow with invertible 1x1 convolutions,'' \textit{Advances in neural information processing systems}, vol. 31, 2018.

\bibitem{srflow}
A. Lugmayr, M. Danelljan, L. Van Gool, R. Timofte, ``Srflow: Learning the super-resolution space with normalizing flow,'' in \textit{Computer Vision--ECCV 2020: 16th European Conference, Glasgow, UK, August 23--28, 2020, Proceedings, Part V 16}, pp. 715--732, 2020.

\bibitem{process1}
E. T. Ziukelis, E. Mak, M. -E. Dounavi, L. Su, J. T. O'Brien, ``Fractal dimension of the brain in neurodegenerative disease and dementia: A systematic review,'' \textit{Ageing Research Reviews}, vol. 79, pp. 101651, 2022.

\bibitem{milv}
T. Stoji{\'c}, I. Reljin, B. Reljin, ``Adaptation of multifractal analysis to segmentation of microcalcifications in digital mammograms,'' \textit{Physica A: Statistical Mechanics and its Applications}, vol. 367, pp. 494--508, 2006.

\bibitem{chongfenxing1}
Y. Xu, H. Ji, C. Ferm{\"u}ller, ``Viewpoint invariant texture description using fractal analysis,'' \textit{International Journal of Computer Vision}, vol. 83, no. 1, pp. 85--100, 2009.

\bibitem{chongfenxing2}
P. Li, Q. Pan, S. Jiang, M. Yan, J. Yan, G. Ning, ``Development of novel fractal method for characterizing the distribution of blood flow in multi-scale vascular tree,'' \textit{Frontiers in Physiology}, vol. 12, pp. 711247, 2021.

\bibitem{fenxing3}
L. G. Souza Fran{\c{c}}a, J. G. Vivas Miranda, M. Leite, N. K. Sharma, M. C. Walker, L. Lemieux, Y. Wang, ``Fractal and multifractal properties of electrographic recordings of human brain activity: toward its use as a signal feature for machine learning in clinical applications,'' \textit{Frontiers in physiology}, vol. 9, pp. 1767, 2018.

\bibitem{fenxing4}
T. Stoji{\'c}, I. Reljin, B. Reljin, ``Adaptation of multifractal analysis to segmentation of microcalcifications in digital mammograms,'' \textit{Physica A: Statistical Mechanics and its Applications}, vol. 367, pp. 494--508, 2006.

\bibitem{hossdoff1}
K. J. Falconer, \textit{The geometry of fractal sets}, no. 85, 1985.

\bibitem{bubianxing}
K. Falconer, \textit{Fractal geometry: mathematical foundations and applications}, 2004.

\bibitem{encoding}
Y. Xu, F. Li, Z. Chen, J. Liang, Y. Quan, ``Encoding spatial distribution of convolutional features for texture representation,'' \textit{Advances in Neural Information Processing Systems}, vol. 34, pp. 22732--22744, 2021.

\bibitem{influnece}
Y. Xu, F. Li, Z. Chen, J. Liang, Y. Quan, ``Multifractal geometry,'' \textit{In Fractal Geometry and Stochastics II}, pp. 3--37, 2000.

\bibitem{error}
Y. Xu, Y. Quan, Z. Zhang, H. Ling, H. Ji, ``Classifying dynamic textures via spatiotemporal fractal analysis,'' \textit{Pattern Recognition}, vol. 48, no. 10, pp. 3239--3248, 2015.

\bibitem{safm}
L. Sun, J. Dong, J. Tang, J. Pan, ``Spatially-Adaptive Feature Modulation for Efficient Image Super-Resolution,'' \textit{arXiv preprint arXiv:2302.13800}, 2023.

\bibitem{fft}
P. Podulka, ``Fast Fourier Transform detection and reduction of high-frequency errors from the results of surface topography profile measurements of honed textures,'' \textit{Eksploatacja i Niezawodno{\'s}{\'c}}, vol. 23, no. 1, pp. 84--93, 2021.

\bibitem{medsegdiff}
J. Wu, R. Fu, H. Fang, Y. Zhang, Y. Yang, H. Xiong, H. Liu, Y. Xu, ``Medsegdiff: Medical image segmentation with diffusion probabilistic model,'' \textit{arXiv preprint arXiv:2211.00611}, 2022.

\bibitem{FFHQ}
T. Karras, S. Laine, T. Aila, ``A style-based generator architecture for generative adversarial networks,'' in \textit{Proceedings of the IEEE/CVF conference on computer vision and pattern recognition}, pp. 4401--4410, 2019.

\bibitem{celeba}
Z. Liu, P. Luo, X. Wang, X. Tang, ``Deep learning face attributes in the wild,'' in \textit{Proceedings of the IEEE international conference on computer vision}, pp. 3730--3738, 2015.

\bibitem{df2k}
E. Agustsson, R. Timofte, ``Ntire 2017 challenge on single image super-resolution: Dataset and study,'' in \textit{Proceedings of the IEEE conference on computer vision and pattern recognition workshops}, pp. 126--135, 2017.

\bibitem{urban}
J. -B. Huang, A. Singh, N. Ahuja, ``Single image super-resolution from transformed self-exemplars,'' in \textit{Proceedings of the IEEE conference on computer vision and pattern recognition}, pp. 5197--5206, 2015.

\bibitem{adam}
D. P. Kingma, J. Ba, ``Adam: A method for stochastic optimization,'' \textit{arXiv preprint arXiv:1412.6980}, 2014.

\bibitem{psnr}
Z. Wang, A. C. Bovik, H. R. Sheikh, E. P. Simoncelli, ``Image quality assessment: from error visibility to structural similarity,'' \textit{IEEE transactions on image processing}, vol. 13, no. 4, pp. 600--612, 2004.

\bibitem{fid}
M. Heusel, H. Ramsauer, T. Unterthiner, B. Nessler, S. Hochreiter, ``Gans trained by a two time-scale update rule converge to a local nash equilibrium,'' \textit{Advances in neural information processing systems}, vol. 30, 2017.

\bibitem{brgm}
R. V. Marinescu, D. Moyer, P. Golland, ``Bayesian image reconstruction using deep generative models,'' \textit{arXiv preprint arXiv:2012.04567}, 2020.

\bibitem{pulse}
S. Menon, A. Damian, S. Hu, N. Ravi, C. Rudin, ``Pulse: Self-supervised photo upsampling via latent space exploration of generative models,'' in \textit{Proceedings of the ieee/cvf conference on computer vision and pattern recognition}, pp. 2437--2445, 2020.

\bibitem{fd_sr_hybrid}
G. Pandey, U. Ghanekar, ``A hybrid single image super-resolution technique using fractal interpolation and convolutional neural network,'' \textit{Pattern Recognition and Image Analysis}, pp. 18--23, 2021.

\bibitem{fd_sr_wee2010novel}
Y. C. Wee, H. J. Shin, ``A novel fast fractal super resolution technique,'' \textit{IEEE Transactions on Consumer Electronics}, vol. 56, no. 3, pp. 1537--1541, 2010.

\bibitem{fd_sr_shao2021noisy}
K. Shao, Q. Fan, Y. Zhang, F. Bao, C. Zhang, ``Noisy single image super-resolution based on local fractal feature analysis,'' \textit{IEEE Access}, vol. 9, pp. 33385--33395, 2021.

\bibitem{fd_sr_xu2013single}
H. Teng Xu, G. Tao Zhai, X. Kang Yang, ``Single image super-resolution with detail enhancement based on local fractal analysis of gradient,'' \textit{IEEE Transactions on circuits and systems for video technology}, vol. 23, no. 10, pp. 1740--1754, 2013.

\bibitem{LSCIDMR}
Bai, Cong, Zhang, Minjing, Zhang, Jinglin, Zheng, Jianwei, and Chen, Shengyong. {\it{LSCIDMR: Large-Scale Satellite Cloud Image Database for Meteorological Research}}. \textit{IEEE Transactions on Cybernetics}, vol. PP, no. 99, pp. 1--13, 2021.

\bibitem{Manga109}
Fujimoto, Azuma, Ogawa, Toru, Yamamoto, Kazuyoshi, Matsui, Yusuke, and Aizawa, Kiyoharu. {\it{Manga109 dataset and creation of metadata}}. In \textit{the 1st International Workshop}, 2016.

\bibitem{set5}
Bevilacqua, Marco, Roumy, Aline, Guillemot, Christine, and Morel, Alberi. {\it{Low-Complexity Single Image Super-Resolution Based on Nonnegative Neighbor Embedding}}. In \textit{British Machine Vision Conference}, 2012.

\bibitem{set14}
Zeyde, Roman, Elad, Michael, and Protter, Matan. {\it{On Single Image Scale-Up Using Sparse-Representations}}. In \textit{International Conference on Curves and Surfaces}, 2010.

\bibitem{LPIPS}
Richard Zhang, Phillip Isola, Alexei A Efros, Eli Shechtman, and Oliver Wang. {\it{The unreasonable effectiveness of deep features as a perceptual metric}}. In \textit{Proceedings of the IEEE/CVF Conference on Computer Vision and Pattern Recognition (CVPR)}, pages 586–595, 2018.

\bibitem{MUSIQ}
Junjie Ke, Qifei Wang, Yilin Wang, Peyman Milanfar, and Feng Yang. {\it{Musiq: Multi-scale image quality transformer}}. In \textit{Proceedings of the IEEE/CVF International Conference on Computer Vision (ICCV)}, pages 5148–5157, 2021.

\bibitem{CLIPIQA}
Jianyi Wang, Kelvin CK Chan, and Chen Change Loy. {\it{Exploring clip for assesses the look and feel of images}}. In \textit{Proceedings of the AAAI Conference on Artificial Intelligence}, 2023.

\bibitem{resnet}
K. He, X. Zhang, S. Ren, and J. Sun, ``Deep residual learning for image recognition,'' in \textit{Proceedings of the IEEE conference on computer vision and pattern recognition}, 2016, pp. 770--778.

\bibitem{alex}
A. Krizhevsky, I. Sutskever, and G. E. Hinton, ``Imagenet classification with deep convolutional neural networks,'' \textit{Advances in neural information processing systems}, vol. 25, 2012.

\bibitem{resshift}
Yue, Zongsheng, Wang, Jianyi, and Loy, Chen Change. {\it{Resshift: Efficient diffusion model for image super - resolution by residual shifting}}. \textit{Advances in Neural Information Processing Systems}, vol. 36, 2024.
\bibitem{diffbir}
Lin, Xinqi, He, Jingwen, Chen, Ziyan, Lyu, Zhaoyang, Dai, Bo, Yu, Fanghua, Qiao, Yu, Ouyang, Wanli, and Dong, Chao. {\it{Diffbir: Toward blind image restoration with generative diffusion prior}}. In \textit{European Conference on Computer Vision}, pages 430 - 448. Springer, 2025.
\bibitem{pasd}
Yang, Tao, Wu, Rongyuan, Ren, Peiran, Xie, Xuansong, and Zhang, Lei. {\it{Pixel - aware stable diffusion for realistic image super - resolution and personalized stylization}}. In \textit{European Conference on Computer Vision}, pages 74 - 91. Springer, 2025.

\bibitem{stablediffusion}
Rombach, Robin, Blattmann, Andreas, Lorenz, Dominik, Esser, Patrick, and Ommer, Björn. {\it{High - resolution image synthesis with latent diffusion models}}. In \textit{Proceedings of the IEEE/CVF conference on computer vision and pattern recognition}, pages 10684 - 10695, 2022.














\end{thebibliography}
\end{document}